\documentclass{article}
\PassOptionsToPackage{hyphens}{url}

\usepackage[preprint]{corl_2026}
\usepackage{graphicx}
\usepackage{booktabs}
\usepackage{subcaption}
\usepackage{float}

\title{Robot-Matched Soft Fingers for UMI Data Collection: Lessons from 100 Hours of Pick and Place}

\author{
  Kobi Kelemen$^{*}$ \quad
  Tomasz Niedziałkowski$^{*}$ \quad
    Marcin Paszczak \quad
  Victoriano Montesinos \\[6pt]
  {\normalfont Lute} \\
  {\normalfont\texttt{\{kornel, tomasz, marcin, victor\}@lute.one}} \\[6pt]
  {\normalfont\small $^{*}$Equal Contribution}
}

\begin{document}
\maketitle

\begin{figure}[H]
  \centering
  \def\teaserh{0.22\linewidth}
  \subcaptionbox{UMI gripper\label{fig:teaser:gripper}}{%
    \includegraphics[height=\teaserh]{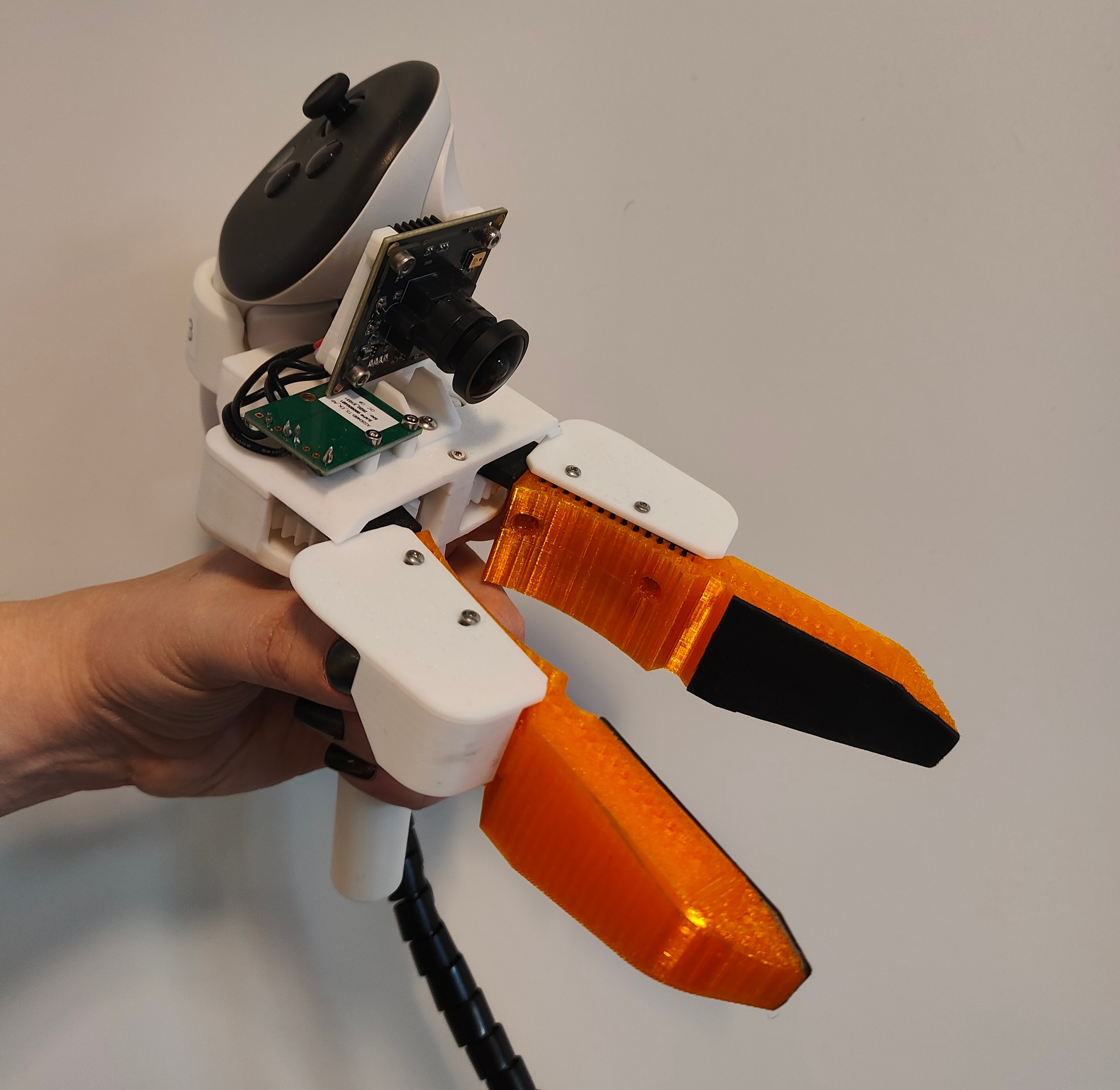}}%
  \hfill
  \subcaptionbox{Data collection\label{fig:teaser:collection}}{%
    \includegraphics[height=\teaserh]{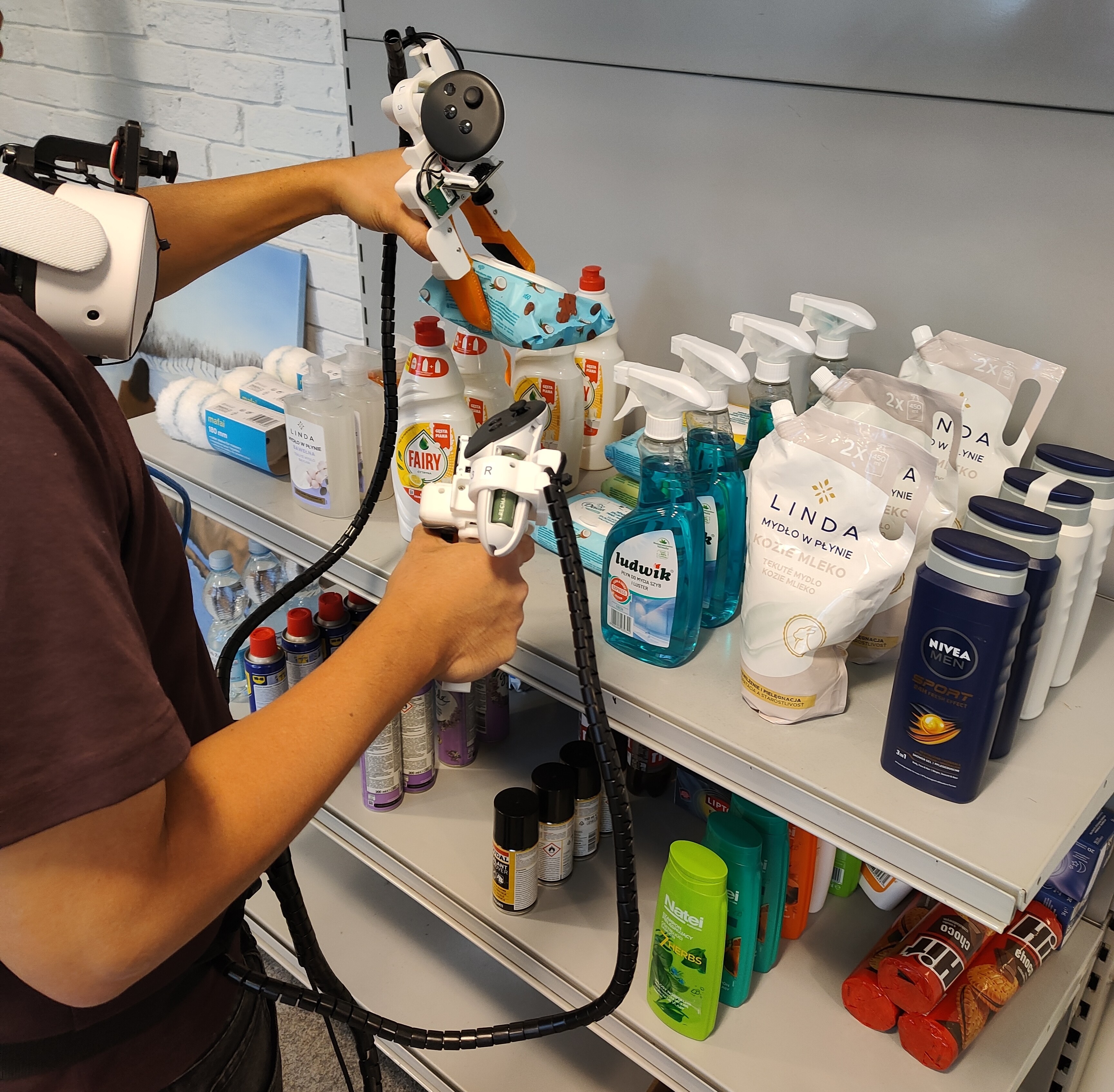}}%
  \hfill
  \subcaptionbox{R1 Lite\label{fig:teaser:lite}}{%
    \includegraphics[height=\teaserh]{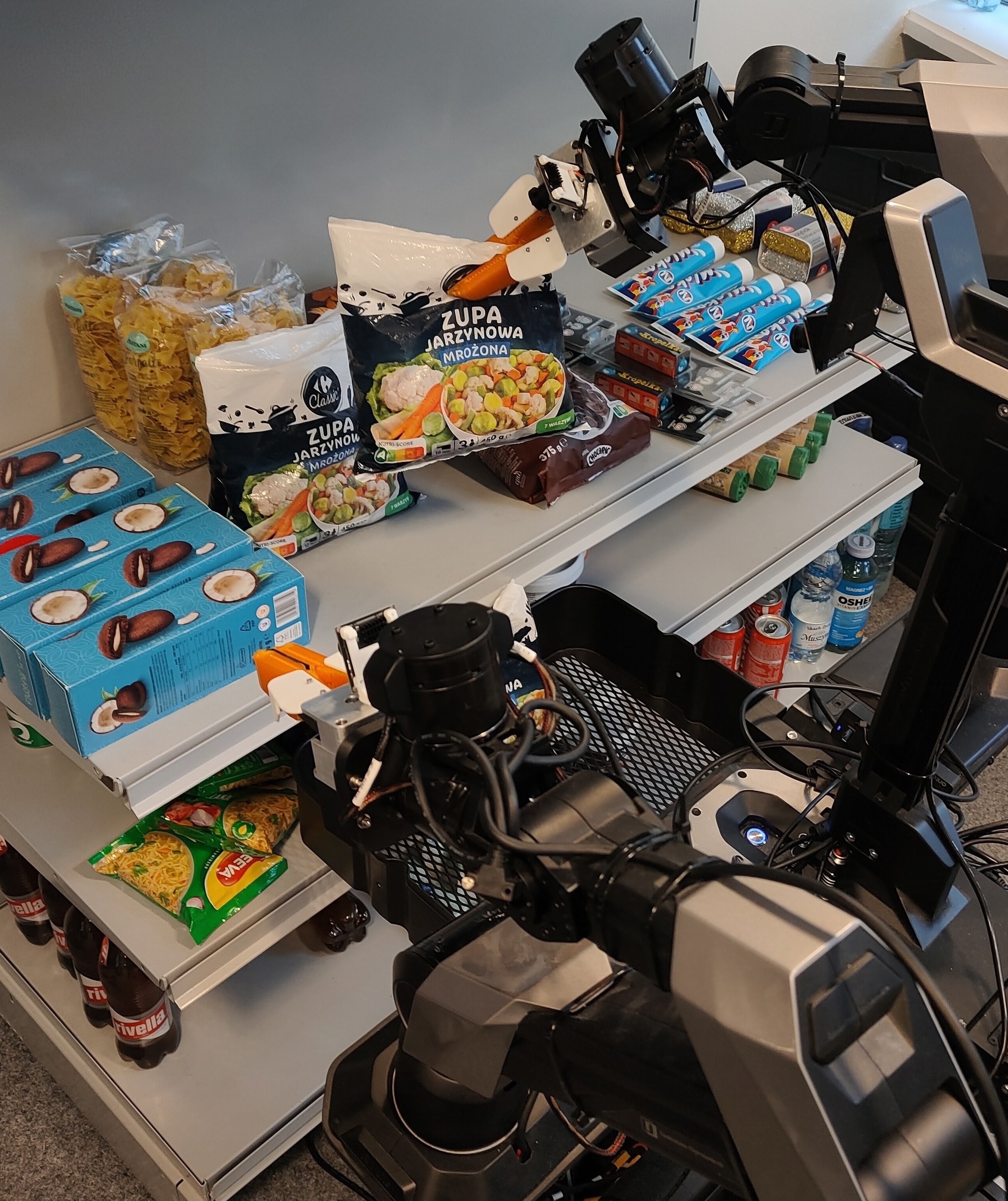}}%
  \hfill
  \subcaptionbox{R1 Pro\label{fig:teaser:pro}}{%
    \includegraphics[height=\teaserh]{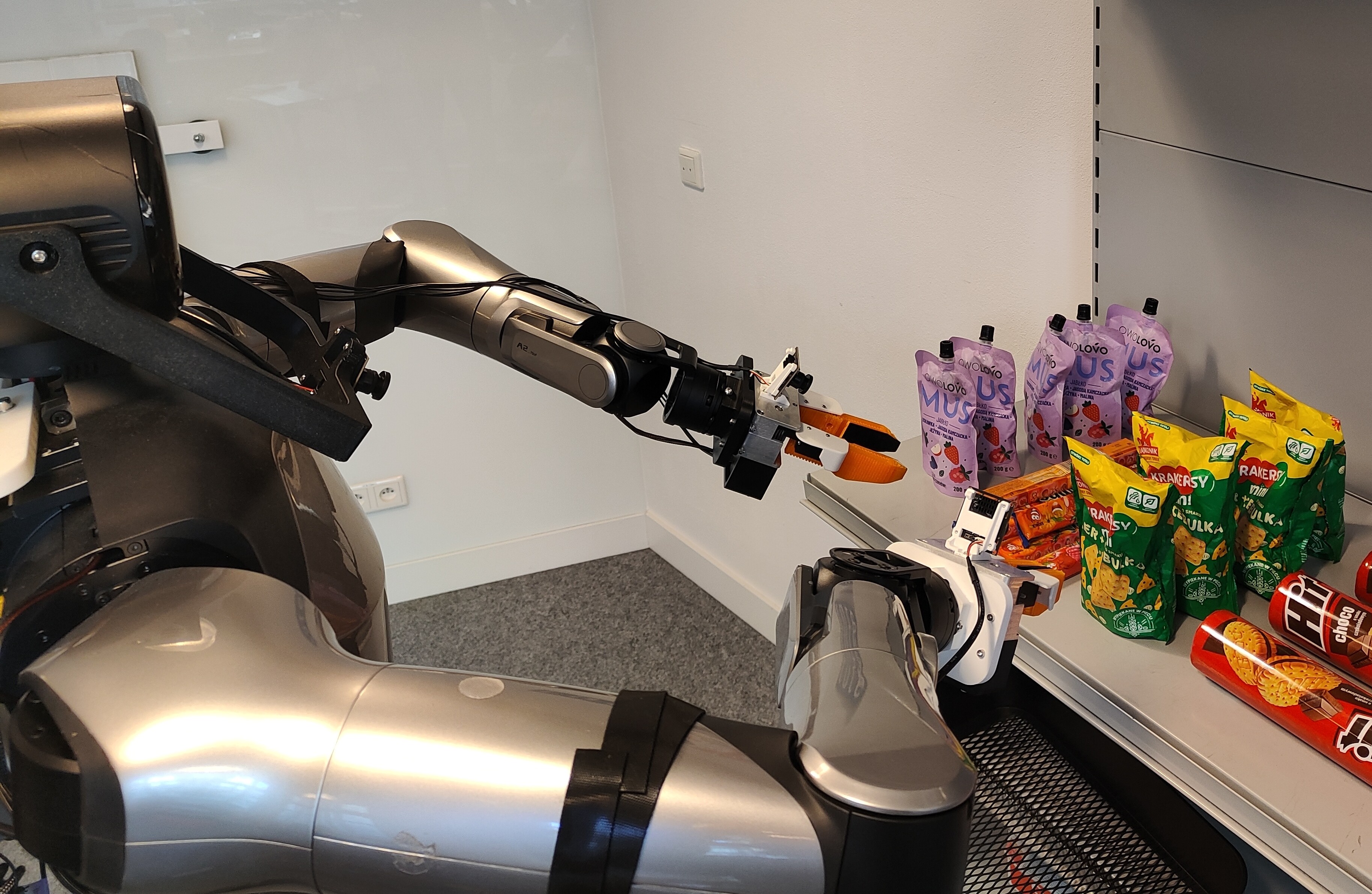}}
  \caption{Our interface is a portable data collection system for multi-task pick and place.
    (a)~Our UMI gripper.
    (b)~Operators collect demonstrations with handheld grippers whose soft, angled fingers match those of the robot,
    battery-powered so it can be used anywhere. Since demonstrations record end-effector poses,
    our policies deploy on any robot with matching grippers that can reach those poses,
    e.g., the Galaxea (c)~R1 Lite and (d)~R1 Pro.}
  \label{fig:teaser}
\end{figure}

\begin{abstract}
Training robot policies for multi-task pick and place requires demonstrations on many objects, which handheld Universal Manipulation Interfaces (UMIs) can collect quickly and cheaply. However, existing UMIs depend on external base stations, use rigid fingers that give no signal about contact forces, or use parallel-jaw fingers that limit grip strength. We present a battery-powered handheld interface, built on YUBI, whose soft, angled fingers wrap around objects, deform visibly under load as a visual proxy for force, and match the grippers of the robot. With it, we collect a 99.7-hour dataset over 53 objects, train $\pi_{0.5}$ policies on 49 of them, and hold out the other 4 as unseen objects. We find that a top camera raises success by 13 points even when only about half of the data includes it, and that mid-training on our multi-task data raises success after fine-tuning on unseen objects from 35\% to 57\%. Finger design, dataset, and data processing code: \href{https://lute-corp.github.io/soft-yubi-site/}{lute-corp.github.io/soft-yubi-site}
\end{abstract}

\keywords{UMI, Soft Grippers, Imitation Learning, Cross-Embodiment Transfer}

\section{Introduction}

General robot policies for multi-task pick and place must handle a wide variety of objects, so they need to be trained on demonstrations with many of them.

The Universal Manipulation Interface (UMI)~\citep{chi2024universalmanipulationinterfaceinthewild} enables collecting such multi-task datasets quickly and cheaply, without a robot in the loop. Follow-up work mostly changes how the handheld gripper is tracked: with a tracking camera~\citep{zhaxizhuoma2025fastumiscalablehardwareindependentuniversal}, a phone~\citep{yin2026feasiblecaprealtimeembodimentconstraint}, a 360$^\circ$ camera~\citep{yang2026kinematicinterfacewildmodular}, head-mounted SLAM~\citep{ai2026hifiumilearningdeployablemanipulation}, self-tracking trackers~\citep{nai2026humanoidmanipulationinterfacehumanoid}, or VR controllers~\citep{ohkawa2026yubiyieldinguniversalbidigital}, while EgoMI~\citep{yu2026egomilearningactivevision} also captures the operator's head motion. However, these interfaces either require external base stations~\citep{cheng2026tacumimultimodaluniversalmanipulation}, use rigid fingers that give the policy no signal about contact forces~\citep{ohkawa2026yubiyieldinguniversalbidigital,ai2026hifiumilearningdeployablemanipulation}, or use parallel-jaw fingers~\citep{chi2024universalmanipulationinterfaceinthewild,zhaxizhuoma2025fastumiscalablehardwareindependentuniversal,yin2026feasiblecaprealtimeembodimentconstraint,nai2026humanoidmanipulationinterfacehumanoid} that touch the object at only two surfaces, which limits grip strength and the range of graspable shapes and breaks the one-to-one mapping to robots with angled grippers~\citep{zhaxizhuoma2025fastumiscalablehardwareindependentuniversal}. KIWI~\citep{yang2026kinematicinterfacewildmodular} also uses deformable fingers, but evaluates only capture: its grippers do not match a robot gripper, and it trains no policies.

To address these limitations, we introduce a portable data collection system (Fig.~\ref{fig:teaser}) built on YUBI~\citep{ohkawa2026yubiyieldinguniversalbidigital}. It needs no base stations and runs from a portable battery and laptop, so it can be used anywhere. Its soft, angled fingers wrap around objects, match our robots' grippers, and deform visibly under load, giving the policy a visual proxy for force.

Our contributions are (1) a portable handheld data collection interface for pick and place, built on YUBI with robot-matched soft fingers; (2) a 99.7-hour dataset over 53 objects, to be released with our finger design and simplified data processing code; and (3) an evaluation of the value of a top camera and of how multi-task mid-training helps policies learn unseen objects.

\section{Data Collection System}
\label{sec:method}

Our interface builds on the YUBI hardware and software stack~\citep{ohkawa2026yubiyieldinguniversalbidigital}: we keep its Meta Quest pose tracking, aperture encoders, and recording software, and describe only our changes.

\subsection{Hardware}
\label{sec:hardware}

\paragraph{UMI gripper.}
Operators collect demonstrations with a UMI gripper based on YUBI, keeping its finger-driven actuation and ESP32-C6 aperture-encoder readout. We replace the rigid fingertips with 1.5$\times$ longer soft TPU ones that open up to 90$^\circ$, conform to more object shapes, and deform visibly under load, a visual proxy for grasp force. For easier sourcing, we swap YUBI's AS5601 encoder for an AS5048A and use a different fisheye camera (Arducam UC684, Appendix~\ref{app:data}).

\paragraph{Robot gripper.}
We build a 450\,g aluminum robot gripper with steel gears, driven by YUBI's Dynamixel XM430-W350-R servo. It shares the UMI gripper's dimensions, fingertips, and camera placement, so demonstrations map one-to-one, and its metal build is durable.

\paragraph{Sensing and power.}
Each UMI gripper carries a wrist camera and a Meta Quest 3S controller tracked by the headset, which the operator wears around the neck with a top camera on it. A portable power station runs it for about 19 hours per charge (Appendix~\ref{app:data}).

\subsection{State and Action Representation}
\label{sec:state-and-action-representation}

For each gripper, our interface records its 6-DoF pose and aperture, giving a 20-dimensional bimanual end-effector vector per frame: position (3), 6D rotation (6), and aperture (1) per arm. The poses are expressed in the Quest headset's tracking frame, whose origin and yaw are arbitrary and change between recordings. Absolute poses in this frame do not transfer to the robot, so we only use quantities that do not depend on its origin or yaw. The frame is gravity-aligned, so each gripper's roll and pitch remain observable.

\paragraph{State.}
Let $T_L, T_R \in SE(3)$ be the current gripper poses, with rotations $R_L, R_R$. The state contains the pose of the left gripper in the frame of the right one, $T_R^{-1} T_L$, as a position and a 6D rotation~\citep{zhou2019continuity}, both apertures $g_L, g_R \in [0, 1]$ (0 closed, 1 open), and the gravity direction in each gripper frame, $R_L^\top z$ and $R_R^\top z$, where $z$ is the up axis of the tracking frame: 17 dimensions in total. The relative pose alone cannot express a tilt shared by both grippers, since rotating both together leaves $T_R^{-1} T_L$ unchanged; the gravity terms recover each gripper's roll and pitch. Unlike EgoMI~\citep{yu2026egomilearningactivevision}, which keeps one hand's pose in a frame aligned at each episode's first timestep, our state contains no absolute pose.

\paragraph{Action.}
The policy predicts chunks of $H = 30$ future gripper poses and apertures. Each future pose $T_{a,h}$ of an arm is expressed in the frame of that arm's current pose $T_s$ as $T_{\mathrm{rel},h} = T_s^{-1} T_{a,h}$, while apertures stay absolute. As in UMI~\citep{chi2024universalmanipulationinterfaceinthewild}, the anchor $T_s$ is re-measured at every chunk rather than fixed at the episode start, whose pose has no counterpart at deployment. Because no input or output depends on the tracking frame's origin or yaw, no calibration between the tracking frame and the robot is needed: at deployment, the predicted chunk is mapped back to absolute targets around the robot's current end-effector pose as $T_s T_{\mathrm{rel},h}$. Appendix~\ref{app:inference} describes their execution.

\subsection{Data Processing Pipeline}
\label{sec:data-processing-pipeline}

Like YUBI, we convert recordings into 30\,Hz LeRobot~\citep{cadene2024lerobot} episodes, with our own conversion, which maps each pose to the gripper's tool center point and rejects episodes in which a sensor stream stalls, keeping 92.03 of the 100.99 hours recorded for the mid-training objects (Appendix~\ref{app:data}).

\section{Experimental Results}
\label{sec:result}

Because our data records end-effector poses, it transfers to any robot with matching grippers whose workspace covers the demonstrations. We use two such robots, a Galaxea R1 Pro and a Galaxea R1 Lite, but only the R1 Pro can reach most of our UMI data (Appendix~\ref{app:ik}), so we evaluate all policies trained or fine-tuned on UMI data on the R1 Pro and use the R1 Lite only for policies fine-tuned on its own teleoperation data. Our dataset contains 99.7 hours of demonstrations (over 51{,}000 episodes) over 53 objects, collected in-house, where each task is picking and placing one everyday object. We mid-train on 92 hours over 49 objects, collected by 5 operators, and hold out the other 4 objects (7.7 hours) as unseen objects (Sec.~\ref{sec:generalization}). We mid-train the $\pi_{0.5}$~\citep{intelligence2025pi05visionlanguageactionmodelopenworld} base checkpoint, pretrained by Physical Intelligence, on our dataset and then fine-tune it on a single object with supervised fine-tuning (SFT), using the relative state with absolute roll and pitch. We run 50 trials per policy and object and report the success rate. Appendices~\ref{app:hyperparameters}--\ref{app:evaluation} give the training, robot, and evaluation details.

\subsection{Does a Top Camera Help?}
\label{sec:top-camera}

\begin{table}[t]
  \centering
  \small
  \caption{Success rate of the mid-trained policy on 5 in-distribution objects, on the R1 Pro.}
  \label{tab:top-camera}
  \begin{tabular}{lccc}
    \toprule
     & Top view in training & \multicolumn{2}{c}{Success rate} \\
    \cmidrule(lr){3-4}
    Object & (\% of episodes) & With top view & Without top view \\
    \midrule
    Peas & 100\% & 78\% & 54\% \\
    Fruit puree & 100\% & 62\% & 52\% \\
    Dried mango & 80\% & 72\% & 62\% \\
    Almonds & 34\% & 70\% & 64\% \\
    Bottle & 0\% & 18\% & 2\% \\
    \midrule
    Mean & -- & 60\% & 47\% \\
    \bottomrule
  \end{tabular}
\end{table}

We test whether a top camera helps beyond the two wrist cameras, to fix the camera setup for Sec.~\ref{sec:generalization}. Only 48 of our 92 hours, covering 30 of the 49 objects at least in part, include a top camera. We mid-train a single policy that uses the top view where it was recorded and a black image elsewhere, and that also blacks out the top view with probability 0.5 per training sample so that it does not depend on it (Appendix~\ref{app:hyperparameters}). We then evaluate this policy with and without the top view, without task-specific fine-tuning. On 5 in-distribution objects on the R1 Pro, the top camera raises the mean success rate from 47\% to 60\% (Table~\ref{tab:top-camera}). The gain is not confined to objects whose own episodes include the view: the bottle has none and still improves. Both settings fail most often on the bottle, which the policy grasps but does not hold. We use both in the rest of our experiments.

\subsection{Does Multi-Task UMI Data Help Learn Unseen Objects?}
\label{sec:generalization}

\begin{figure}[t]
  \centering
  \includegraphics[width=\linewidth]{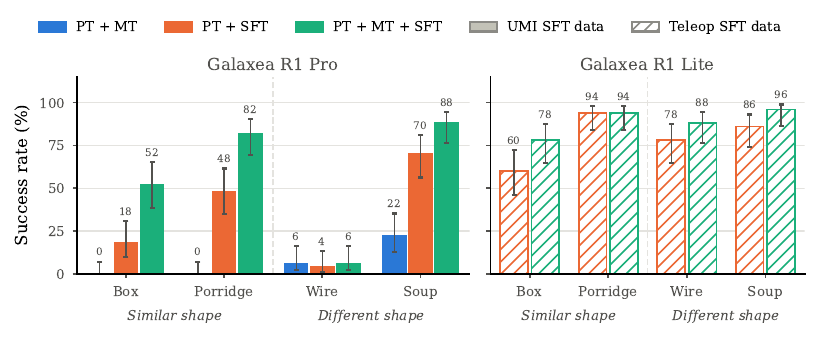}
  \caption{Success rate on each out-of-distribution (OOD) object, grouped by whether its shape is similar to or different from those in the mid-training data. Left: on the R1 Pro, we compare mid-training and SFT on UMI data. Right: on the R1 Lite, we compare the same policies fine-tuned on teleoperation data (hatched). Error bars show 95\% Wilson intervals.}
  \label{fig:generalization}
\end{figure}

We collect SFT data for 4 objects outside the mid-training data (OOD): a box and a porridge, with shapes similar to the mid-training objects, and a wire and a soup, with different shapes. Starting from the $\pi_{0.5}$ base checkpoint (PT), we compare (i) the mid-trained policy without task-specific data (PT + MT, R1 Pro only), (ii) PT fine-tuned on one object (PT + SFT), and (iii) the mid-trained policy fine-tuned on it (PT + MT + SFT), all with the wrist and top cameras. SFT uses 1.5 to 2.4 hours of data per object (Appendix~\ref{app:hyperparameters}): UMI data on the R1 Pro, and teleoperation data on the R1 Lite, which tests whether mid-training on UMI data still helps on a robot that cannot execute most of its motions. On the R1 Pro, mid-training raises success after SFT from 35\% to 57\% (Fig.~\ref{fig:generalization}): from 45\% to 74\% on the box, porridge, and soup, which tolerate imprecise grasps, while both policies fail on the wire, which needs a precise grasp (4\% and 6\%). On the R1 Lite, it raises success from 80\% to 89\%, although the policy fine-tuned without it already performs well. Mid-training on diverse UMI data therefore helps the policy learn new objects that tolerate imprecise grasps, even on the R1 Lite, but does not yet enable precise manipulation. Without task-specific data, PT + MT succeeds in only 7\% of the trials, so at our data scale some SFT data remains necessary.

\section{Discussion and Limitations}
\label{sec:discussion}

Our experiments suggest two lessons. First, a top camera is worth adding even when only part of the data includes it: trained with top-view dropout, the policy works without the view and gains 13 points with it (Sec.~\ref{sec:top-camera}). Second, diverse multi-task UMI data pays off when learning new objects: mid-training on it raised success after SFT from 35\% to 57\% on the R1 Pro, mostly on objects that tolerate imprecise grasps, and still helped on the R1 Lite, which cannot reach most of the UMI data (Sec.~\ref{sec:generalization}). The benefit likely comes from what the policy learns about objects and tasks, not from the motions.

\paragraph{Limitations.}
The handheld devices are cabled to the battery and laptop, limiting the operator's reach, and inside-out tracking degrades when the grippers leave the headset cameras' view. Demonstrations recorded without reachability feedback may not transfer to robots with a smaller workspace, such as the R1 Lite; real-time feedback during collection~\citep{yin2026feasiblecaprealtimeembodimentconstraint} could address this. Our evaluation covers pick and place in one environment on a few objects, so single-object differences are noisy and we pool success rates, and we chose the state representation from preliminary tests rather than a full ablation. Our policies also struggle with tasks needing precise grasps, such as the wire, and often do not close the gripper tightly enough, so the bottle slips out. These are our main focus for future work, with tactile sensing and broader evaluation.

\acknowledgments{We thank Marcin Gendek, Gracjan Góral, Jan Wawszczak, Wojciech Krupiński, Krzysztof Walas, and Paweł Budzianowski for their contributions to our robotics and AI stack, and Valeriia Shkurko, Aksana Matsiusheuskaya, Teng Lai, Gabriela Myrcha, and Katarzyna Rećko for collecting the dataset.}

\bibliography{main}

\section*{Author Contributions}
\textbf{Kobi Kelemen} implemented the data processing and inference pipeline, led data collection and evaluations, trained the policies, and contributed to project ideation and to writing this report.

\textbf{Tomasz Niedziałkowski} designed and built the hardware, and contributed to code reviews and writing this report.

\textbf{Marcin Paszczak} designed and built the initial hardware.

\textbf{Victoriano Montesinos} supervised the project from its inception to implementation and evaluation, contributed to project ideation and code reviews, and led the experiment design and the writing of this report.

\appendix

\section{Data Collection Details}
\label{app:data}

\paragraph{Sensing and power.}
Each UMI gripper carries a wrist camera and a Meta Quest 3S controller tracked by the headset's inside-out cameras. A top camera mounted on the headset provides the view used in Secs.~\ref{sec:top-camera} and~\ref{sec:generalization}. All cameras are Arducam UC684 boards (115$^\circ$ HFOV, 640$\times$480 at 30\,Hz). The operator wears the headset around the neck, not on YUBI's chest belt. A 1024\,Wh EcoFlow DELTA~3 powers the headset, USB hubs, and an HP ProBook 465 laptop directly over USB, for about 19 hours per charge. A full set costs about 3{,}000\,USD.

\paragraph{Data processing.}
Like YUBI, we convert recordings into 30\,Hz LeRobot~\citep{cadene2024lerobot} episodes and filter out defective ones, but with our own conversion. We resample each recording onto a uniform 30\,Hz grid with zero-order hold, using the recorder's log time because the wrist cameras' header stamps are about 0.5\,s off, and its one annotated segment becomes one episode. We move each tracked pose to the tool center point where the finger pads meet, using a fixed transform from the gripper model, so the pose does not depend on the aperture, which we normalize to $[0, 1]$; the action at time $t$ is the state at $t+1$. Missing or unused camera views are replaced with a black image at both training and inference time. Because zero-order hold hides dropped messages, we store a per-frame freshness flag for every stream and reject an episode if any stream repeats its value for three or more consecutive frames (about 0.1\,s), or if fewer than 95\% of its pose and gripper frames are fresh. Of the 100.99 hours recorded for the mid-training objects, this filtered 8.9\%, leaving 92.03 hours. States and actions are z-score normalized over all datasets.

\section{Model Hyperparameters}
\label{app:hyperparameters}

All runs train $\pi_{0.5}$ on 8 NVIDIA B200 GPUs with a batch size of 1024 and a constant learning rate of $2 \times 10^{-4}$ after 1k warmup steps, using the relative state with absolute roll and pitch, full relative actions, and top-view dropout with probability 0.5. This state representation worked better than the alternatives we tried in preliminary tests; since the best one may depend on the pretrained model, a systematic comparison is left for future work. The top camera is present in 55 of the 113 mid-training datasets, and a dropout probability of 0.5 is high enough that every object is often seen without the top view but low enough that the policy still learns from it. We do not evaluate PT without fine-tuning, since its action space differs from ours. Mid-training runs on our multi-task dataset, and we use its checkpoint after 80k steps for all evaluations of the mid-trained policy (Secs.~\ref{sec:top-camera} and~\ref{sec:generalization}) and to initialize SFT. SFT starts either from the $\pi_{0.5}$ base checkpoint (PT + SFT) or from this mid-training checkpoint (PT + MT + SFT), with the same data for both, and we evaluate every SFT checkpoint after 4k steps. Table~\ref{tab:sft} lists the SFT data per object.

\begin{table}[h]
  \centering
  \small
  \caption{SFT data per object, in hours.}
  \label{tab:sft}
  \begin{tabular}{llcc}
    \toprule
    Object & Shape & UMI (R1 Pro) & Teleop (R1 Lite) \\
    \midrule
    Box & Similar & 1.93 & 1.45 \\
    Porridge & Similar & 1.86 & 1.85 \\
    Wire & Different & 1.58 & 1.67 \\
    Soup & Different & 2.32 & 2.37 \\
    \bottomrule
  \end{tabular}
\end{table}

\section{Inference Setup}
\label{app:inference}

\paragraph{Robots.}
The R1 Pro has two 7-DoF arms and a 4-DoF torso, and the R1 Lite has the stock 6-DoF Galaxea A1 arms and a 3-DoF torso. Both observe the scene through their wrist cameras and a top camera, which sits where the operator wears the headset: on the R1 Pro's chest, where it moves with the torso like the headset with the operator, and on a fixed mount on the R1 Lite. We collect the R1 Lite teleoperation data with GELLO~\citep{wu2023gello}. The policy is trained at 30\,Hz and predicts chunks of 30 steps, which we execute at 30\,Hz on the R1 Lite and at 20\,Hz on the R1 Pro, where the slower motion keeps the gripper from being damaged when it hits the shelf; in our tests, the execution rate did not change the success rate.

The robot and the policy server communicate over gRPC. The robot sends observations and receives action chunks asynchronously, so its control loop never blocks on inference; if no chunk arrives within 1\,s, the request is reissued. The server maps the predicted relative actions back to absolute end-effector targets around the current end-effector pose (Sec.~\ref{sec:state-and-action-representation}), and the robot computes that pose from its measured joints by forward kinematics. The policy runs on a single NVIDIA GeForce RTX 5080 GPU in a PC in the same room as the robot, connected over wired Ethernet through a local router, while the control loop runs on the robot.

\paragraph{Chunk execution.}
We use real-time chunking (RTC)~\citep{black2025realtimeexecutionactionchunking}, with the implementation from LeRobot~\citep{cadene2024lerobot}. The robot requests the next chunk when 15 of the 30 actions remain, sending the unexecuted actions along with it, and skips the first actions of the new chunk that the measured inference latency has already covered. The robot therefore keeps moving while the next chunk is computed. On the server, RTC uses an execution horizon of 20 steps, a maximum guidance weight of 20, and a linear prefix-attention schedule.

\paragraph{R1 Lite.}
A 30\,Hz control loop executes one action per tick. For each action, PyRoki~\citep{kim2025pyroki} solves inverse kinematics for both 6-DoF arms toward the gripper tips, warm-started from and regularized toward the previous solution; the 3-DoF torso stays fixed, since our software did not yet support moving it. The joint positions go to the arm driver, which runs at 200\,Hz with velocity and acceleration limits. Each gripper's Dynamixel servo maps the aperture linearly onto its calibrated travel in current-based position mode, which caps the grip force. The run stops if no action arrives for 2\,s.

\paragraph{R1 Pro.}
For each action, we solve a single inverse kinematics problem with PyRoki over both 7-DoF arms and all four torso joints, targeting both gripper tips in the chassis frame. A cost keeps the solution close to the previous one, more strongly for the torso, and the torso joints are limited to $\pm 20^\circ$ around their start pose. A 20\,Hz control loop executes one action per tick, and the joint targets are linearly interpolated and sent to Galaxea's motion controller, which rejects arm steps above $45^\circ$ and torso steps above $0.3^\circ$ per command, and the grippers are driven as on the R1 Lite.

\section{Inverse Kinematics Reachability Analysis}
\label{app:ik}

To check whether a robot can execute our UMI data, we solve inverse kinematics for every frame and measure the position error between the target and the reached gripper tip. We call an episode unreachable when the 95th percentile of this error exceeds a threshold. To place the recorded poses on the robot, we anchor them at a representative start pose sampled from the training data.

\begin{figure}[h]
  \centering
  \includegraphics[width=\linewidth]{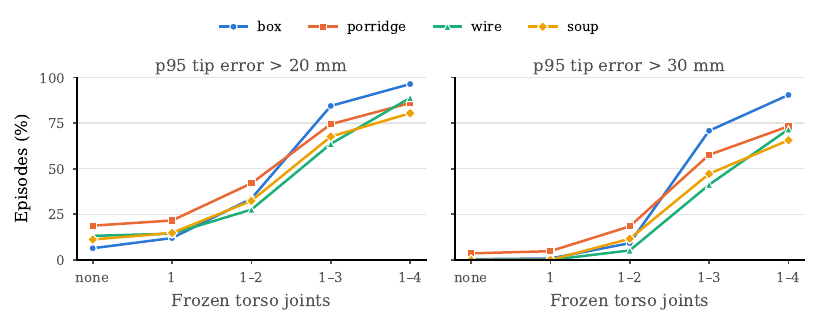}
  \caption{Share of R1 Pro episodes whose 95th-percentile IK tip error exceeds 20\,mm (left) or 30\,mm (right), as torso joints are frozen from the chassis (joint 1) up to the arms (joint 4).}
  \label{fig:ik-reachability}
\end{figure}

With whole-body inverse kinematics, the R1 Pro reaches most of the UMI episodes we collected for SFT on the four unseen objects: 12\% exceed 20\,mm and 1\% exceed 30\,mm (Fig.~\ref{fig:ik-reachability}). The share is highest for the porridge (19\%) and the wire (13\%), where the operator reaches to the back of the shelf. Freezing torso joints shrinks the workspace quickly: with the two lowest joints frozen, 34\% of the episodes exceed 20\,mm, and with three, 73\%. The torso is therefore essential for executing our data. The R1 Lite, whose 3-DoF torso we keep fixed in our evaluations, cannot reach most of our data: on the mid-training data, 83\% of the episodes have a 95th-percentile error above 20\,mm (69\% above 30\,mm).

We use this analysis to choose which robot to evaluate each policy on. We do not filter the mid-training data by reachability, nor the SFT data for the box, porridge, and soup. For the wire, we drop SFT episodes whose 99th-percentile tip error exceeds 20\,mm, which removes 18\% of its 1.93 hours (Table~\ref{tab:sft} lists the 1.58 hours that remain).

\section{Evaluation Protocol}
\label{app:evaluation}

We use the same protocol for all experiments. During evaluation of each object, we use random background objects on the shelf. Each trial has a 30\,s time limit and is a success if the policy places the object into the basket located between the robot and the shelf within the limit, and a failure otherwise. The policy may retry the grasp within the limit, and a human operator judges success. We place a row of the target objects directly in front of the robot, and do not replace the objects from the basket unless a failure occurs or until the row of objects is complete. After each row replacement we shuffle the objects by 2\,cm backwards up to 6\,cm. The robot distance to the shelf is kept constant across rollouts and we use a fixed start pose from the training dataset.
\end{document}